\documentclass{article}

\usepackage{microtype}
\usepackage{graphicx}
\usepackage{subcaption}
\usepackage{svg}
\usepackage{booktabs} 

\usepackage{caption}
\usepackage{subcaption}

\usepackage{hyperref}
\usepackage{xcolor}

\usepackage[preprint]{icml2026}

\usepackage{amsmath}
\usepackage{amssymb}
\usepackage{mathtools}
\usepackage{amsthm}
\usepackage{multirow}
\usepackage{pifont}

\usepackage[capitalize,noabbrev]{cleveref}

\theoremstyle{plain}

\theoremstyle{definition}

\theoremstyle{remark}

\usepackage[textsize=tiny]{todonotes}

\icmltitlerunning{Language-Assisted Image Clustering Guided by Discriminative Relational Signals and Adaptive Semantic Centers}
\begin{document}

\twocolumn[
  \icmltitle{Language-Assisted Image Clustering Guided by Discriminative Relational Signals and Adaptive Semantic Centers}




  \begin{icmlauthorlist}
    \icmlauthor{Jun Ma}{seu}
    \icmlauthor{Xu Zhang}{seu}
    \icmlauthor{Zhengxing Jiao}{seu}
    \icmlauthor{Yaxin Hou}{seu}
    \icmlauthor{Hui Liu}{sfu}
    \icmlauthor{Junhui Hou}{cityu}
    \icmlauthor{Yuheng Jia}{seu}


  \end{icmlauthorlist}

  \icmlaffiliation{seu}{School of Computer Science and Engineering, Southeast University, Nanjing, China}
  \icmlaffiliation{sfu}{School of Computing Information Sciences, Saint Francis University, Hong Kong, China}
  \icmlaffiliation{cityu}{Department of Computer Science, City University of Hong Kong, Hong Kong, China}
 
  \icmlcorrespondingauthor{Yuheng Jia}{yhjia@seu.edu.cn}

  \icmlkeywords{Machine Learning, ICML}

  \vskip 0.3in
]



\printAffiliationsAndNotice{}  

\begin{abstract}
Language-Assisted Image Clustering (LAIC) augments the input images with additional texts with the help of vision-language models (VLMs) to promote clustering performance. Despite recent progress, existing LAIC methods often overlook two issues: (i) textual features constructed for each image are highly similar, leading to weak inter-class discriminability; (ii) the clustering step is restricted to pre-built image-text alignments, limiting the potential for better utilization of the text modality. To address these issues, we propose a new LAIC framework with two complementary components. First, we exploit cross-modal relations to produce more discriminative self-supervision signals for clustering, as it compatible with most VLMs training mechanisms.
Second, we learn category-wise continuous semantic centers via prompt learning to produce the final clustering assignments.
Extensive experiments on eight benchmark datasets demonstrate that our method achieves an average improvement of \textbf{2.6\%} over state-of-the-art methods, and the learned semantic centers exhibit strong interpretability. \textbf{ Code is available in the supplementary material.}
\end{abstract}

\section{Introduction} 
Image clustering aims to automatically partition images into
semantically coherent clusters without any supervision.
To handle the complex structure of high-dimensional natural images, deep clustering methods~\cite{SCAN2020, Spice2022, ProPos2022, lhy2025, CDC2025} utilize deep neural networks for image clustering.
Recently, by leveraging the strong image-text alignment capability of vision-language models (VLMs, e.g., CLIP \cite{CLIP2021}), \textbf{Language-Assisted Image Clustering (LAIC)}~\cite{SIC2023, MCA2024, TAC2024, GNC2025} has been proposed, which incorporates textual semantics to enhance clustering performance, as visually similar categories may be far apart in linguistic space. 
Existing LAICs methods can be roughly summarized into two steps. (i) \textbf{Text counterpart construction}. They extract a candidate noun set from external corpus (e.g., WordNet \cite{wordnet1995} to describe images of interest, and construct a text feature for each sample based on several most similar nouns\footnote{In practice, some methods use similarity to weight the features of all nouns in the candidate set. However, after applying the softmax function with a small temperature parameter, almost all of the weights approach zero, and only the top-k most similar nouns determine the text features constructed for each sample.}. Then self-supervised signals such as positive image-text pairs or pseudo-labels can be derived. (ii) \textbf{Clustering with images and texts}. They leverage features of both image and text modalities to perform clustering (e.g., by K-means or training clustering heads).

\begin{figure*}[t]
    \centering
    \includegraphics[width=\textwidth, height=10cm,keepaspectratio]{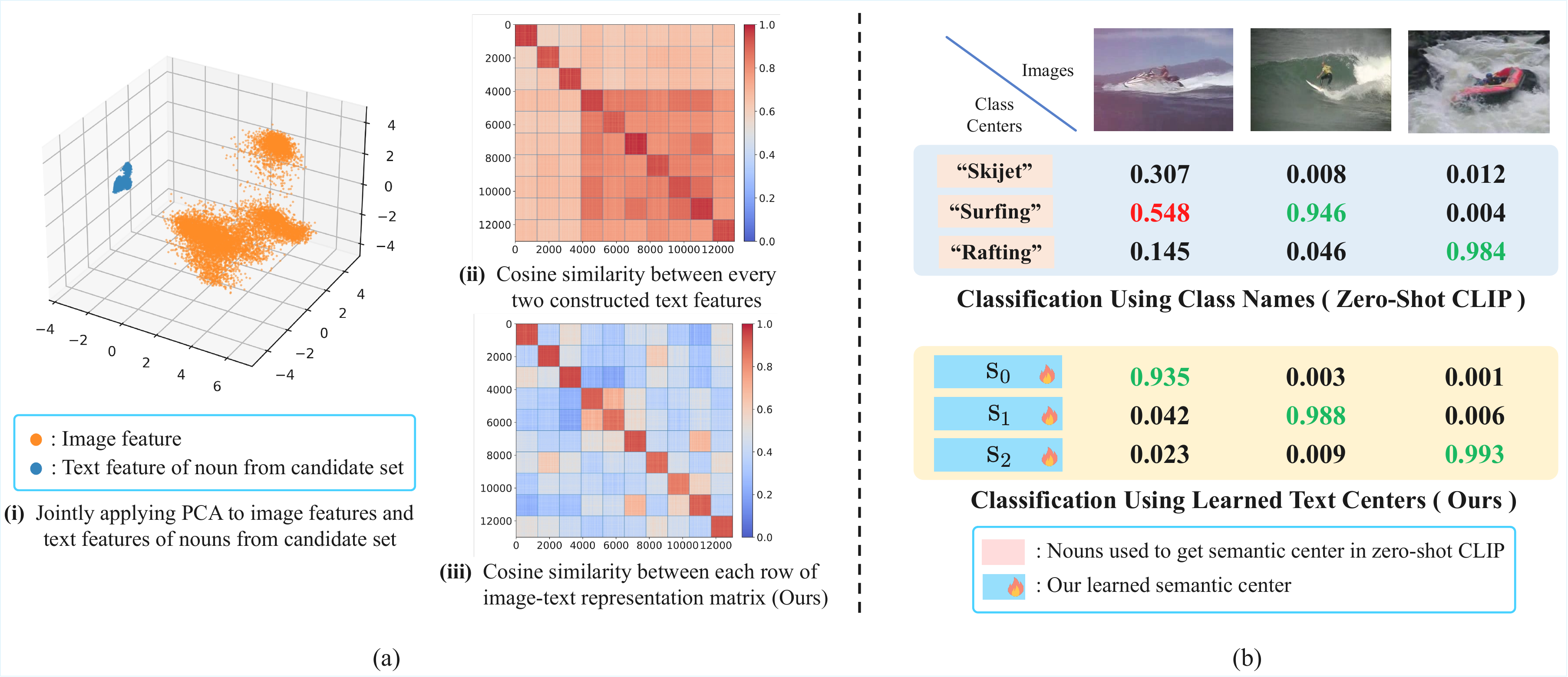}

  \caption{
  \textbf{(a)} Analysis of the phenomena in the first step of existing LAICs and our method on the ImageNet-10 dataset. 
  (i) Jointly PCA results shows text features of noun from candidate set exhibit a more compact distribution, which are closed to each other compared to image features.
  (ii) Due to the phenomenon in (i), the similarity of text features across different samples are overall highly similar, leading to weak inter-class discriminability. 
  (iii) Rows of our learned image-text representation matrix performs as new representation, showing better inter-class discriminability with in-class consistency retained. 
  \textbf{(b)} Classification scores of three images (from DTD dataset) with respect to different semantic centers obtained by different methods. Zero-shot CLIP relies on semantic centers using ground-truth class names, which may not always capture accurate semantics (e.g, an image of motorboat is incorrectly assigned to ``Surfing”. Our learned semantic center can be even better than class names, showing stronger discriminability.}
  \label{fig:1}
\end{figure*}

Despite encouraging progress, we find that existing methods suffer from two main problems.  
(i) \textit{During the first step}, the text features constructed for samples are overall highly similar, which largely affects the discriminability of the supervision signals derived from the text modality (shown in Fig.~\ref{fig:1}(a-ii)).
This likely occurs because the text prompts used for the text encoder typically capture only coarse, high-level semantics, whereas images contain rich visual details and fine-grained information (e.g., an image of a cat playing on grass under a blue sky with clouds may be described simply as “a photo of a cat”).
As a result, the text features of nouns from the candidate set are very close to each other than those of images (shown in Fig.~\ref{fig:1}(a-i)), leading to similar text features constructed for each sample. 
(ii) \textit{During the second step}, the current methods train the clustering model on existing manually constructed semantic space, which is constrained by previously established image-text alignments. This approach restricts the possibility of further extracting more accurate semantics from the native semantic space of VLMs.

To address these issues, we propose a new LAIC framework. 
First, to enhance the discriminability of the text modality, we construct an image-text representation matrix by minimizing the discrepancy between the two modalities. 
Each row of this matrix characterizes how an image can be represented from the perspective of the textual space, which can thus be treated as a new representation of this image. 
This approach of extracting cross-modal signals is \textbf{compatible with the training mechanisms of many VLMs }such as CLIP, which are pretrained by aligning image-text pairs with similar semantics. 
Therefore, these new representations substantially improve inter-class discriminability and maintain intra-class compactness (shown in Fig.~\ref{fig:1}(a-iii)). We then directly perform K-means clustering on the representation matrix to obtain a strong baseline. 

Second, to move beyond the constraints of the constructed image-text matching relationships during clustering, we construct learnable category-wise semantic centers via prompt learning from the native semantic space of VLMs. 
Specifically, we initialize semantic centers by prompting the text encoder with the template ``a photo of a [class]”, where [class] is replaced by a learnable vector for each pseudo-label class.  
We optimize these centers by maximizing the alignment between each pseudo-label class's semantic center feature and image features. 
The final clustering is achieved by assigning each sample to the closest semantic center. 
It is worth noting that, unlike the approaches in model adaptation where the prompt prefix is learned with class names available, we fix the prefix and optimize the semantic centers.
Our approach closely mirrors CLIP's zero-shot process, and the results show that we can even learn better semantic centers than those constructed using the true class names (shown in Fig.~\ref{fig:1}(b)), suggesting that these continuous centers can provide more accurate semantics for each class in view of CLIP. Meanwhile, the learned semantic centers exhibit strong interpretability.

Our contributions are summarized as follows:
\begin{itemize}
    \item We propose a novel cross-modal relational mining which is highly compatible with the mechanisms of VLMs. By leveraging the inherent connections between image and text modalities, the constructed representations for samples are better in inter-class discriminability, providing richer self-supervision signals and a strong baseline for clustering.
    \item We introduce a continuous category-wise semantic centers learning strategy by prompt learning directly in the semantic space of VLMs. This approach moves beyond the constraints of the pre-built image-text alignments and acquires more accurate semantic centers, allowing more effective utilization of the text modality. 
    \item Experimental results on eight public datasets demonstrate that our method achieves an average improvement of 2.6\% over state-of-the-art methods, and the learned semantic centers exhibit strong interpretability.
\end{itemize}

\section{Related Work}
\subsection{Deep Image Clustering}
Traditional image clustering mainly relies on K-means~\cite{kmeans1967} and Spectral Clustering~\cite{spectral2007}, which struggle to handle high-dimensional data. Methods such as DEC~\cite{DEC2016} and DeepCluster~\cite{DeepClustering2018} introduce representation learning into the clustering pipeline, marking the rise of deep clustering. Approaches including SCAN~\cite{SCAN2020} and SPICE~\cite{Spice2022} leverage pseudo-labeling to further improve clustering by selecting high-confidence pseudo-labels from the obtained cluster assignments. Meanwhile, contrastive learning contributes to notable progress in image clustering. CC~\cite{CC2021} and TCL~\cite{TCL2022} perform contrastive learning at both instance and cluster levels, while ProPos~\cite{ProPos2022} combines non-contrastive learning at the instance level with contrastive learning at the cluster level. Some methods analyze phenomena observed in clustering from different perspectives. CDC~\cite{CDC2025} identifies the overconfidence problem in deep clustering and alleviates it by introducing a model structure with calibration head. DCBoost~\cite{lhy2025} researchs the inconsistency between global and local feature structures in deep clustering, and proposes a plugin that enhances global features by leveraging local structural information. With the advent of vision-language models (VLMs, e.g., CLIP~\cite{CLIP2021}), methods such as TEMI~\cite{TEMI2023} and CPP~\cite{CPP2024} exploit the strong representations learned by VLMs to enhance clustering performance. 

\subsection{Language-Assisted Image Clustering}
Language-Assisted image clustering incorporates supervision signals extracted from the text modality to enhance image clustering performance, as textual semantics can provide additional guidance, especially when the categories are visually similar but semantically different. SIC~\cite{SIC2023} derives pseudo-label supervision according to relationships between images and semantics to guide image clustering, while MCA~\cite{MCA2024} proposes a hierarchy-based nouns filtering strategy and improves image-text alignments at three levels. TAC~\cite{TAC2024} treats signals from text modality as external guidance, enhancing clustering performance via self-distillation between two modalities. GradNorm~\cite{GNC2025} selects more accurate texts that are more semantically aligned with images from the gradient perspective. 
With the help of text criteria and large language models (LLMs), more challenging multiple clustering tasks can be enhanced. IC$|$TC~\cite{ictc} proposes a new method for image clustering based on user-specified text criteria, allowing users to have significant control over the clustering results. Multi-Sub~\cite{multisub} aligns user preferences expressed through textual prompts with visual representations by leveraging the synergistic capabilities of CLIP and LLMs.

\section{Method}
\paragraph{Overview.} 

Given a set of unlabeled images $\mathcal{X}=\{x_i\}_{i=1}^{N}$, image clustering aims to partition these samples into semantically coherent clusters, where only the cluster number is known.
To this end, we propose a new LAIC framework built upon CLIP with two components. First, we design a cross-modal relation mining scheme to extract discriminative self-supervision signals from text modality in Sec.~\ref{sec:3.1}. Second, we learn continuous semantic centers for each class from CLIP's semantic space by prompt learning in Sec.~\ref{sec:3.2}. These semantic centers serve as discriminative anchors, thereby producing the final clustering results.

\subsection{Semantic Space Construction and Supervision Mining}
\label{sec:3.1}

\paragraph{Constructing  dataset-specific semantic space.}
We adopt CLIP as the backbone and denote its image encoder and text encoder as $f(\cdot)$ and $g(\cdot)$, respectively. 
We extract image feature $\mathbf{x}_i$ for each $x_i \in \mathcal{X}$ using $f(\cdot)$ and form the image feature matrix 
$\mathbf{X} = [\mathbf{x}_1,\dots,\mathbf{x}_N]^\top \in \mathbb{R}^{N\times d}$. Meanwhile, we encode each word $w_i$ from external corpus WordNet~\cite{wordnet1995} into a text feature $\mathbf{w}_i$ using $g(\cdot)$ and form the text feature matrix $\mathbf{W} = [\mathbf{w}_1,\dots,\mathbf{w}_L]^\top \in \mathbb{R}^{L\times d},$
where $L$ is the size of WordNet. 

To obtain a condidate nouns set that can semantically describe the current dataset, we follow TAC~\cite{TAC2024} and first compute fine-grained semantic centers of images. Specifically, we run K-means clustering on $\mathbf{X}$ to produce $\tilde{k}$ clusters. We set $\tilde{k}=\left\lceil N/300\right\rceil$; for datasets where the number of samples in each ground-truth class is smaller than $300$, we use $\tilde{k}=3K$, where $K$ is the cluster number.  
Let $\{ \mathcal{P}_r\}_{r=1}^{\tilde{k}}$ be the resulting partition. The $r$-th fine-grained semantic center is computed as the mean feature:
\begin{equation}
\mathbf{p}_r = \frac{1}{|\mathcal{P}_r|}\sum_{i\in \mathcal{P}_r}\mathbf{x}_i,\quad r=1,\dots,\tilde{k}.
\label{eq:center}
\end{equation}

Next, each text feature $\mathbf{w}_i$ from $\mathbf{W}$ is assigned to its nearest image center $\mathbf{p}_{a(i)}$, where
\begin{equation}
a(i)=\arg\max_{r\in\{1,\dots,\tilde{k}\}} \ \mathrm{cos}(\mathbf{w}_i,\mathbf{p}_r),
\quad i=1,\dots,L,
\label{eq:assign_word}
\end{equation}
and
\begin{equation}
    \cos(a, b) = \frac{a^\top b}{\|a\|_2 \|b\|_2}
    \label{eq:cosine_sim}
\end{equation}
means cosine similarity.

Then, for each center $\mathbf{p}_r$, we select $\theta$ most similar text features among those assigned to it, noted as $\mathbf{U}_r$.
Finally, we take the union $\bigcup_{r=1}^{\tilde{k}}\mathbf{U}_r$ as the dataset-specific text features candidate set $\mathbf{U} = [\mathbf{u}_1,\dots,\mathbf{u}_M]^\top \in \mathbb{R}^{M\times d}$, where $M=|\mathbf{U}|$.

\paragraph{Mining discriminative supervision via cross-modal relational signals extracting.}
Existing LAIC methods typically retrieve text feature for each image based on the several nearest text features in $\mathbf{U}$ and build supervision. However, as illustrated in Fig.\ref{fig:1}(a-ii), this strategy overlooks that the text features constructed for samples are highly similar overall.
Moreover, it ignores abundant discriminative information carried by \emph{non-neighbor} texts, as samples belonging to the same semantic typically exhibit consistent dissimilarity with the same irrelevant texts.

To address these issues, we propose to mine supervision from a cross-modal perspective by learning an image-text representation matrix $\mathbf{C}\in\mathbb{R}^{N\times M}$. It reconstructs image features using the entire candidate set $\mathbf{U}$:
\begin{equation}
\min_{\mathbf{C}} \ \|\mathbf{X}-\mathbf{C}\mathbf{U}\|_F^2 + \gamma \|\mathbf{C}\|_F^2,
\label{eq:ridge}
\end{equation}
where the first term enforces that each image feature can be well approximated by a weighted combination of text features, 
and the Frobenius regularizer imposes a smoothness constraint on $\mathbf{C}$, preventing it from being overly sparse. $\gamma$ is the regularization parameter.

Problem~\eqref{eq:ridge} is a standard ridge regression with a closed-form solution:
\begin{equation}
\mathbf{C}^\star = \mathbf{X}\mathbf{U}^\top \left(\mathbf{U}\mathbf{U}^\top + \gamma \mathbf{I}_M\right)^{-1},
\label{eq:closed_form}
\end{equation}
where $\mathbf{I}_M \in \mathbb{R}^{M \times M}$ represents an identity matrix. Each row $\mathbf{c}_i$ of $\mathbf{C}^\star$ can be interpreted as a dense semantic description of image $x_i$ over the whole candidate noun set. It becomes a discriminative representation for $x_i$, for semantically similar images tend to be compactness in this space, and vice versa. In practice, we directly perform K-means on rows of $\mathbf{C}^\star$ to obtain an initial clustering baseline, which also serves as improved supervision for the subsequent semantic center learning step.

\paragraph{Remark 1.} \textit{Why are cross-modal relational signals more discriminative?} Recall that CLIP aligned a large number of image-text pairs during pretraining stage. Thus, CLIP's discriminative ability across different categories is largely driven by cross-modal information, i.e., the difference between feature similarities of images and texts. The cross-modal relational signals provided by $\mathbf{C}$ align well with this mechanism of CLIP. 

\subsection{Semantic Centers Learning}
\label{sec:3.2}
In this part, we retain only the pseudo-label information from Sec.~\ref{sec:3.1} to overcome the limitations of pre-built image-text alignments. We firstly filter pseudo-labels of high quality and learn semantics for each class to get final clustering assignments. 

\paragraph{High-quality pseudo-labels selection.} Although the previous step provides a strong baseline, the pseudo-labels still contain noise. We therefore filter them to obtain pseudo-labels of high quality with more reliable supervision. To this end, we first define sample neighborhood relations in the space of $\mathbf{C}\in\mathbb{R}^{N\times M}$, where $\mathbf{c}_i$ denotes the $i$-th row of $\mathbf{C}$. We compute $cos(\textbf{c}_i,\textbf{c}_j)$ as similarity between $x_i$ and $x_j$, where $ j\neq i$. For each sample $x_i$, we select the top-$\hat{k}$ most similar samples according to cosine similarity in space of $\textbf{C}$ as its $\hat{k}$-nearest-neighbors ($\hat{k}$-nn) set $\mathcal{N}_{\hat{k}}(i)$.

\begin{figure}
    \centering
    \includegraphics[width=0.9\linewidth]{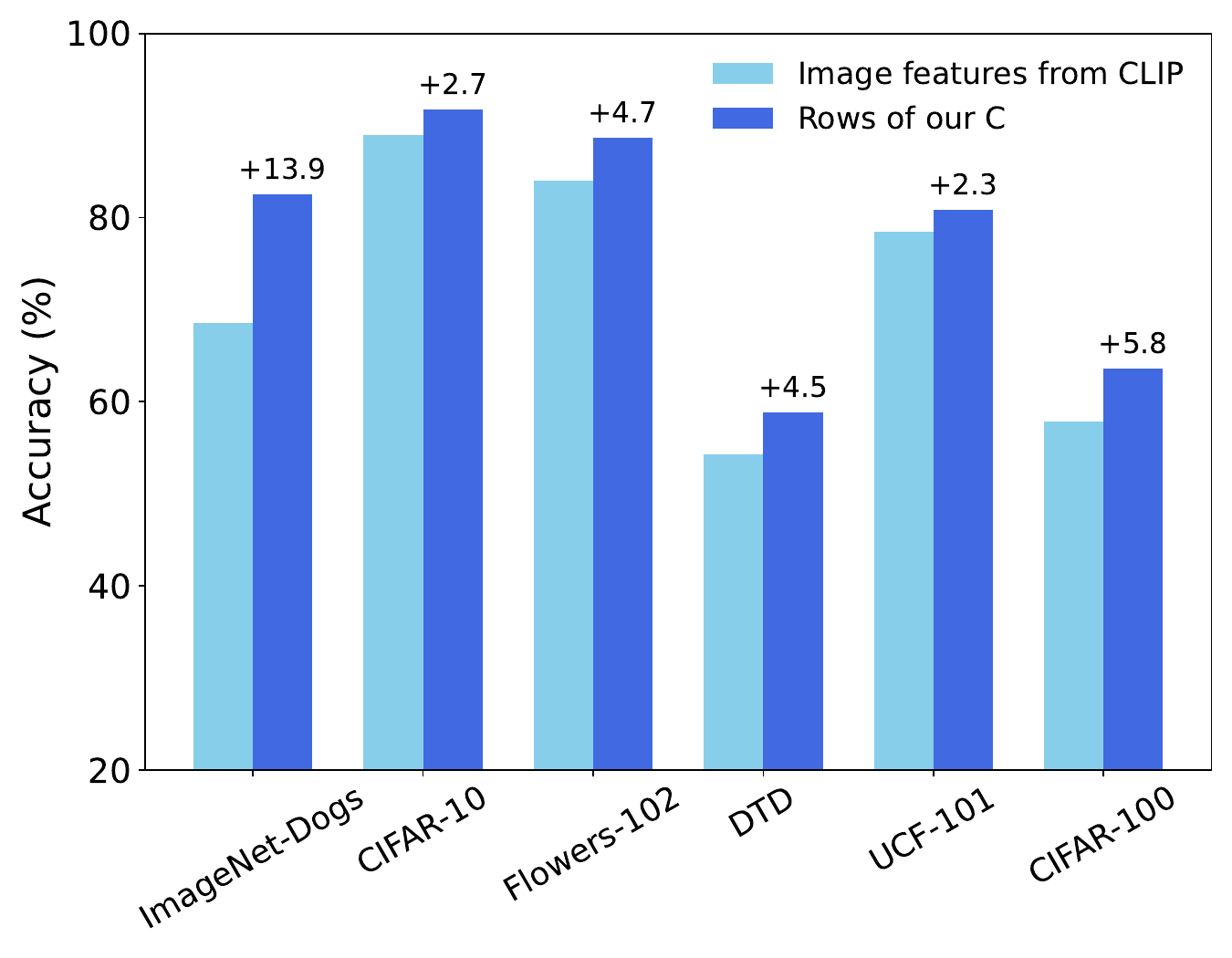}
    \caption{Accuracy of each sample and its $\hat{k}$-nn ($\hat{k}=10$) belonging to the same ground-truth class on different space. We compare the results by computing similarity on images features from CLIP and rows of $\mathbf{C}$, showing our image-text representation matrix has better neighbor consistency.}
    \label{fig:2}
\end{figure}

We observe that a sample and its $\hat{k}$-nn have a high probability of belonging to the same semantic category (shown in Fig.~\ref{fig:2}).
Based on this observation, we hypothesize that \textit{the stronger the pseudo-label consistency between a sample and its neighbors, the more likely the pseudo-label is correct}. Therefore, for $\hat{y}_i\in\{1,\dots,K\}$, we define a neighbor-consistency score:
\begin{equation}
\alpha_i=\frac{1}{\hat{k}}\sum_{j\in\mathcal{N}_{\hat{k}}(i)}\mathbb{I}\big[\hat{y}_j=\hat{y}_i\big],
\end{equation}
where $\mathbb{I}[\cdot]$ denotes the indicator function. Higher neighbor-consistency score means that $\hat{y}_i$ is more likely to be accurate. We select samples with $\alpha_i\ge \tau$ to form the high-quality set
\begin{equation}
\mathcal{D}_L=\{\,(x_i,\hat{y}_i )  \mid \alpha_i\ge \tau \}
\end{equation}
as supervision for semantic centers learning.

\paragraph{Semantic-centers parameterization.} With the pre-trained parameters of the VLMs frozen, we adopt prompt learning to adapt the text-side representations to the target data distribution. We replace the \emph{class} position in the prompt ``a photo of a [class]" with randomly initialized learnable vectors, which serve as the semantic-center variables of pseudo-label classes.


For each class $k$, we introduce a learnable class vectors $\mathbf{v}_k = \big[V^{k}_{1}, V^{k}_{2}, \ldots, V^{k}_{B}\big]$, where each $V^{k}_{b}$ ($b\in\{1,\ldots,B\}$) is a continuous vector in the word embedding space, and $B$ is the number of learnable tokens.
By appending these learnable tokens to fixed prefix tokens, we construct the prompt for class $k$ as 
$\textbf{t}_k = [\mathbf{E}_{prefix},\mathbf{v}_k]$,
where $\mathbf{E}_{prefix}$ contains the embeddings of the fixed prefix tokens, and we set prefix as ``a photo of a''. We feed $\mathbf{t}_k$ into the frozen text encoder $g(\cdot)$ of CLIP to obtain the semantic center's text feature $\mathbf{s}_k$ for class $k$, and $\mathcal{S}=\{ \mathbf{s}_k\}_{k=1}^K$ denotes the set of learnable semantic centers. 


Given an image $x$, we extract its image feature $\mathbf{x} = f(x)$, where $f(\cdot)$ is the frozen CLIP image encoder. 
We define the probability of assigning $x$ to $k$ as 
\begin{equation}
    p(k\mid x)=\frac{\exp(\ell_k(x))}{\sum_{j=1}^{K}\exp(\ell_j(x))},
\end{equation}
where $\ell_k(x)=\cos(\mathbf{x},\mathbf{s}_k)/T$ and $T$ is the temperature parameter learned during pre-training of CLIP.
Thus, the predicted probability distribution of $x$ over $K$ classes is
\begin{equation}
    \mathbf{P}(x)=\big(  \ell_1(x),\ldots,\ell_K(x) \big).
\end{equation}

\paragraph{Semi-supervised training objective.} 
We set $\mathcal{D}_L=\{(x_i,\hat{y}_i) \mid \alpha_i \ge \tau \}$ as the high-quality set, where $\hat{y}_i\in\{1,\ldots,K\}$ as pseudo-labeled set, and  $\mathcal{D}_U=\{x_i \mid \alpha_i < \tau \}$ as the unlabeled set.


Firstly, on $\mathcal{D}_L$, we minimize the generalized cross-entropy (GCE) as supervised loss to enhance the model's robustness against noise in pseudo-labels. 
The loss is formulated as
\begin{equation}
    \mathcal{L}_{\mathrm{sup}}
    = \mathbb{E}_{(x,\hat{y})\sim \mathcal{D}_L}\left[
    \frac{1-\Big(p\big(\hat{y}\mid \mathcal{A}_s(x) \big)\Big)^q}{q}
    \right],
\end{equation}
where $\mathcal{A}_s(\cdot)$ denotes the strong augmentation, $q\in(0,1]$ is the parameter that balances the convergence speed of cross-entropy loss and the noise-robustness of mean absolute error loss. We set $q=0.8$.
This term encourages samples from $\mathcal{D}_L$ to be close to the semantic center of their assigned class guided by pseudo-labels and far from other centers, thus improving inter-class separability.

Next, to provide stable constraints on unlabeled data, we generate weak augmentation for each $x\in\mathcal{D}_U$, denoted as $\mathcal{A}_w(x)$. 
We enforce prediction consistency between the two views by minimizing the mean squared error between the two probability distribution:
\begin{equation}
\mathcal{L}_{\mathrm{con}}
= \mathbb{E}_{x\sim \mathcal{D}_U}\left[\left\|\mathbf{P}\big( \mathcal{A}_s(x) \big)-\mathbf{P}\big( \mathcal{A}_w(x) \big)\right\|_2^2\right].
\end{equation}

Then, to encourage a sufficiently diverse usage of the $K$ centers, we maximize a global entropy regularization:
\begin{equation}
\mathcal{L}_{\mathrm{ent}}
= - \sum_{k=1}^{K} q(k)\log q(k),
\end{equation}

where we define $q(k) = \mathbb{E}_{x\sim (\mathcal{D}_L \cup \mathcal{D}_U)}\big[p(k\mid x)\big]$ as the global average prediction distribution on both high-quality set and unlabeled set.

Finally, we combine the above terms and get the overall objective:
\begin{equation}
\mathcal{L}
= \mathcal{L}_{\mathrm{sup}}
+ \lambda_1\,\mathcal{L}_{\mathrm{con}}
- \lambda_2\,\mathcal{L}_{\mathrm{ent}},
\end{equation}
where weight parameters $\lambda_1 = 2$ and $\lambda_2 = 0.1$ are fixed on all datasets.

\paragraph{Model training and prediction.} We freeze CLIP's image and text encoders and only update $\mathcal{V}=\{\mathbf{v}_k\}_{k=1}^{K}$ using $\mathcal{L}$. 
After training, we obtain $K$ semantic centers $\mathcal{S} = \{\mathbf{s}_k\}_{k=1}^{K}$. For a test image $x$, we extract its image feature $\mathbf{z}=f(x)$ and assign it to the closest semantic center:
\begin{equation}
\tilde{y}(x) = \arg\max_{k}\ \cos(\mathbf{x}, \mathbf{s}_k).
\end{equation}
According to this assignment, we partition images with the same $\tilde{y}$ to the same cluster and get final cluster results. Intuitively, $\mathcal{S}$ provides a set of continuous and discriminative cluster anchors in CLIP's semantic space, enabling stable test-time clustering without relying on a discrete candidate noun set.

\paragraph{Remark 2.} \textit{What’s the difference between prompt tuning in model adaptation and our learning of semantic centers?} In field of model adaptation, the class names are available, and the approaches often replace the prompt prefix before \emph{class} with learnable vectors. The goal is to learn prompts that are better adapted to downstream tasks, in order to better leverage the generalization ability of VLMs. With the unsupervised setting of LAIC, our method fix the prefix and replace \emph{class} with learnable vectors, aiming at learning continuous semantic centers to get clustering results.

\section{Experiments} \label{sec experiments}
In this section, we evaluate our method by comparing it with 18 deep clustering approaches and zero-shot CLIP. We provide a detailed analysis of the experimental results and conduct extensive ablation studies to verify the effectiveness of our method. Additional experimental results and parameter sensitivity analyses are provided in the Appendix~\ref{appendix std}.

\begin{table*}[t]
\centering
\caption{Clustering performance (in percent \%) across four widely used datasets. ZS-CLIP means zero-shot CLIP.}
\label{table main result}
\begin{small}
\setlength{\tabcolsep}{3pt}
\begin{tabular}{l|c|ccc|ccc|ccc|ccc|c}
\toprule
\multicolumn{2}{l|}{Datasets} 
& \multicolumn{3}{c|}{ImageNet-Dogs}
& \multicolumn{3}{c|}{ImageNet-10} 
& \multicolumn{3}{c|}{STL-10}
& \multicolumn{3}{c|}{CIFAR-10}
& \multirow{2}{*}{Avg.}\\
\cmidrule(r){1-14}
Methods & Backbone 
& NMI & ACC & ARI 
& NMI & ACC & ARI 
& NMI & ACC & ARI 
& NMI & ACC & ARI \\
\midrule
DEC \cite{DEC2016}& ResNet 
& 12.2 & 19.5 & 7.9
& 28.2 & 38.1 & 20.3
& 13.6 & 18.5 & 5.0
& 25.7 & 30.1 & 16.1 & 19.6\\

IIC \cite{IIC2019}& ResNet 
& - & - & - 
& - & - & - 
& 49.6 & 59.6 & 39.7
& 51.3 & 61.7 & 41.1 & -\\

DCCM \cite{DCCM2019} & ResNet 
& 32.1 & 38.3 & 18.2
& 60.8 & 71.0 & 55.5 
& 37.6 & 48.2 & 26.2 
& 49.6 & 62.3 & 40.8 & 45.1\\

PICA \cite{PICA2020} & ResNet 
& 33.6 & 32.4 & 17.9
& 78.2 & 85.0 & 73.3 
& 59.2 & 69.3 & 50.4 
& 56.1 & 64.5 & 46.7 & 55.6\\

BYOL \cite{BYOL2020} & ResNet 
& 69.7 & 72.9 & 60.9
& 88.4 & 94.7 & 88.9 
& 75.4 & 86.1 & 71.5 
& 78.0 & 87.5 & 75.2 & 79.1\\

SCAN \cite{SCAN2020} & ResNet 
& 61.2 & 59.3 & 45.7
& - & - & - 
& 69.8 & 80.9 & 64.6 
& 79.7 & 88.3 & 77.2 & - \\


CC \cite{CC2021} & ResNet 
& 44.5 & 42.9 & 27.4
& 85.9 & 89.3 & 82.2 
& 76.4 & 85.0 & 72.6 
& 70.5 & 79.0 & 63.7 & 68.3 \\


MiCE \cite{MICE2021} & ResNet 
& 42.3 & 43.9 & 28.6
& - & - & - 
& 63.5 & 75.2 & 57.5 
& 73.7 & 83.5 & 69.8 & -\\

GCC \cite{GCC2021} & ResNet 
& 49.0 & 52.6 & 36.2
& 84.2 & 90.1 & 82.2 
& 68.4 & 78.8 & 63.1 
& 76.4 & 85.6 & 72.8 & 70.0\\

TCC \cite{TCC2021} & ResNet  
& 55.4 & 59.5 & 41.7
& 84.8 & 89.7 & 82.5 
& 73.2 & 81.4 & 68.9 
& 79.0 & 90.6 & 73.3 & 73.3 \\

TCL \cite{TCL2022} & ResNet 
& 62.3 & 64.4 & 51.6
& 87.5 & 89.5 & 83.7 
& 79.9 & 86.8 & 75.7 
& 81.9 & 88.7 & 78.0 & 77.5\\

DMICC \cite{DMICC2023} & ResNet  
& 58.1 & 58.7 & 43.8
& 91.7 & 96.2 & 91.6 
& 68.9 & 80.0 & 62.5 
& 74.0 & 82.8 & 69.0 & 73.1 \\

LFSS \cite{LFSS2025}& ResNet  
& 61.7 & 69.1 & 53.3
& 85.6 & 93.2 & 85.7 
& 77.1 & 86.1 & 74.0 
& 84.1 & 92.4 & 84.2 & 78.9 \\

\midrule
SIC \cite{SIC2023}& ViT-B/32 
& 69.0 & 69.7 & 55.8
& 97.0 & 98.2 & 96.1 
& 95.3 & 98.1 & 95.9 
& 84.7 & 92.6 & 84.4 & 86.4\\

MCA \cite{MCA2024} & ViT-B/32 
& 73.3 & 74.9 & 61.6
& - & - & - 
& 95.5 & 98.1 & 96.0 
&  \underline{84.9} &  \underline{92.7} &  \underline{84.6} & -\\

TAC \cite{TAC2024} & ViT-B/32 
& 80.6 &  \underline{83.0} &  \underline{72.2}
& 98.5 & 99.2 & 98.3 
& 95.5 & 98.2 & 96.1 
& 83.3 & 91.9 & 83.1 &  \underline{90.0}\\

PRO-DSC \cite{PRODSC2025} & ViT-B/32 
& - & - & - 
& 98.0 & 99.0 & 97.8 
& 95.4 & 98.1 & 95.9 
& 79.6 & 87.1 & 80.2 & -\\

GradNorm \cite{GNC2025} & ViT-B/32 
& \underline{81.0} & 81.2 & 70.9
&  \underline{98.7} &  \underline{99.4} &  \underline{98.7} 
&  \underline{95.6} &  \underline{98.3} &  \underline{96.2} 
& 82.6 & 91.1 & 81.5 & 89.6\\

Ours & ViT-B/32 
& \textbf{86.2} & \textbf{88.0} & \textbf{82.6}
& \textbf{99.6} & \textbf{99.8} & \textbf{99.7} 
& \textbf{96.1} & \textbf{98.5} & \textbf{96.7} 
& \textbf{85.2} & \textbf{92.9} & \textbf{84.8} & \textbf{92.5}\\

\textcolor{gray}{ZS-CLIP \cite{CLIP2021} }& \textcolor{gray}{ViT-B/32}
& \textcolor{gray}{80.6} & \textcolor{gray}{83.0} & \textcolor{gray}{72.2}
& \textcolor{gray}{95.8} & \textcolor{gray}{97.6} & \textcolor{gray}{94.9} 
& \textcolor{gray}{93.9} & \textcolor{gray}{97.1} & \textcolor{gray}{93.7} 
& \textcolor{gray}{80.7} & \textcolor{gray}{90.0} & \textcolor{gray}{79.3} & \textcolor{gray}{88.2} \\
\bottomrule
\end{tabular}
\end{small}
\end{table*}

\vskip 0.15in

\begin{table*}[t]
\centering
\caption{Clustering performance (in percent \%) across four datasets with larger cluster numbers.}
\label{table main result 2}
\begin{small}
\setlength{\tabcolsep}{3pt}
\begin{tabular}{l|c|ccc|ccc|ccc|ccc|c}
\toprule
\multicolumn{2}{l|}{Datasets} & 
\multicolumn{3}{c|}{Flowers-102} &
\multicolumn{3}{c|}{CIFAR-100} & 
\multicolumn{3}{c|}{DTD} & 
\multicolumn{3}{c|}{UCF-101}& 
\multirow{2}{*}{Avg.}\\
\cmidrule(r){1-14}
Methods & Backbone &NMI &ACC &ARI &NMI &ACC &ARI &NMI &ACC &ARI &NMI &ACC &ARI \\
\midrule
SCAN \cite{SCAN2020} & ResNet 
& 77.9 & 56.5 & 50.9
& 55.4 & 38.7 & 25.1
& 59.4 & 46.4 & 31.7 
& 79.7 & 61.1 & 53.1 & 53.0\\
LFSS \cite{LFSS2025} & ResNet   
& 79.2 & 55.4 & 51.8
& 66.0 & 52.4 & 37.6
& 58.7 & 45.5 & 31.9
& 58.9 & 34.5 & 21.6 & 49.5\\
\midrule
SIC \cite{SIC2023}& ViT-B/32 
& 86.9 & 70.2 & 64.2
& 63.5 & 48.3 & 34.7
& 59.6 & 45.9 & 30.5 
& 81.0 & 61.9 & 52.4 & 58.3\\
TAC \cite{TAC2024} & ViT-B/32
&  \underline{87.7} &  \underline{71.8} &  \underline{64.3}
& 68.1 & 55.4 &  \underline{42.4}
& 62.1 & 50.1 &  \underline{34.4} 
& 82.8 &  \underline{68.9} &  \underline{60.6} &  \underline{62.4}\\
PRO-DSC \cite{PRODSC2025} & ViT-B/32 
& 83.2 & 69.8 & 59.6
&  \underline{68.2} &  \underline{55.7} & 40.2  
& 57.6 & 48.1 & 31.3 
& 81.0 & 61.9 & 52.8 & 59.1\\
GradNorm \cite{GNC2025} & ViT-B/32 
& - & - & -
& 67.2 & 54.8 & 37.0
&  \underline{63.1} &  \underline{50.9} & 34.2 
&  \underline{82.9} & 63.2 & 53.9 & - \\
Ours & ViT-B/32 
& \textbf{89.6} & \textbf{78.6} & \textbf{72.9}
& \textbf{68.8} & \textbf{58.1} & \textbf{43.4}
& \textbf{63.8} & \textbf{53.7} & \textbf{36.2} 
& \textbf{84.4} & \textbf{70.2} & \textbf{61.9} & \textbf{65.1}\\

\textcolor{gray}{ZS-CLIP \cite{CLIP2021}}& \textcolor{gray}{ViT-B/32}
& \textcolor{gray}{82.3} & \textcolor{gray}{66.7} & \textcolor{gray}{58.1}
& \textcolor{gray}{67.4} & \textcolor{gray}{60.9} & \textcolor{gray}{38.0}
& \textcolor{gray}{56.5} & \textcolor{gray}{43.1} & \textcolor{gray}{26.9} 
& \textcolor{gray}{79.9} & \textcolor{gray}{63.4} & \textcolor{gray}{50.2} & \textcolor{gray}{57.8}\\ 

\bottomrule
\end{tabular}
\end{small}
\end{table*}

\subsection{Experimental Setup}\label{sec:4.1}
\paragraph{Datasets.} To evaluate the performance of our method, we first apply it to four widely-used image clustering datasets including ImageNet-Dogs~\cite{IN-sub}, ImageNet-10~\cite{IN-sub}, STL-10~\cite{STL10}, CIFAR-10~\cite{CIFAR}, along with four datasets with larger cluster numbers, including Flowers-102~\cite{flowers2}, CIFAR-100~\cite{CIFAR}, DTD~\cite{DTD} and UCF-101~\cite{UCF}. Detailed description of datasets can be found in the Appendix~\ref{appendix datasets}. We train and evaluate our method on the train and test splits, respectively.

\paragraph{Evaluation metrics.} We evaluate clustering performance using three widely adopted metrics: Normalized Mutual Information (NMI), Accuracy (ACC), and Adjusted Rand Index (ARI). NMI and ACC range in $[0,1]$, while ARI ranges in $[-1, 1]$. For all three metrics, higher scores correspond to better clustering performance.

\paragraph{Implementation details. }Following previous works~\cite{SIC2023,TAC2024}, we adopt the pre-trained CLIP model with ViT-B/32~\cite{ViT} and Transformer~\cite{Transformer} as image and text backbones, respectively. We adopt data augmentation methods from~\cite{TAC2024} as weak augmentation $\mathcal{A}_w$, and incorporate RandAugment~\cite{randaug} to construct strong augmentation $\mathcal{A}_s$.
Following~\cite{coop}, training is done for 20 epochs using batch size of 32, with SGD optimizer and an initial learning rate of $2e-3$, which is decayed by the cosine annealing rule. We fix $\theta=2, \gamma=5$, $\hat{k}=10$ and $\tau=1$ in all the experiments. The only exception is that we set $\hat{k}=1$ on datasets with larger cluster numbers, including Flowers-102, CIFAR-100, DTD and UCF-101. All experiments are conducted on a single NVIDIA RTX 4090 GPU.

\subsection{Main Results}
\paragraph{Clustering performance. } We present clustering experimental results on four commonly used datasets in Table~\ref{table main result}, and the results on four datasets with larger cluster numbers in Table~\ref{table main result 2}. The best and second-best results are denoted in bold and underline, respectively. The upper and lower sections in the tables correspond to using ResNet18 (34) and CLIP ViT-B/32 as the backbone, respectively. We primarily focus on comparisons with CLIP-based methods and zero-shot CLIP.

It can be observed that our method consistently outperforms all baseline methods, achieving an average improvement of 2.6\% over the state-of-the-art LAIC method TAC. In particular, our performances improve more on datasets with a larger cluster numbers, with an average increase of 3.5\% in ACC over TAC, indicating that our method performs better on these more complex datasets. Meanwhile, the improvements on fine-grained datasets are particularly significant, such as ImageNet-Dogs (+ 5.0\% in ACC, + 10.4\% in ARI) and Flowers-102 (+ 6.8\% in ACC and + 8.6\% in ARI) over TAC, respectively. We also outperform zero-shot CLIP by an average of 5.8\%, demonstrating that our method learns more accurate semantic centers for different classes.

\begin{figure}[h!]
    \centering
    \begin{subfigure}[b]{0.49\columnwidth}
        \centering
        \includegraphics[width=\linewidth]{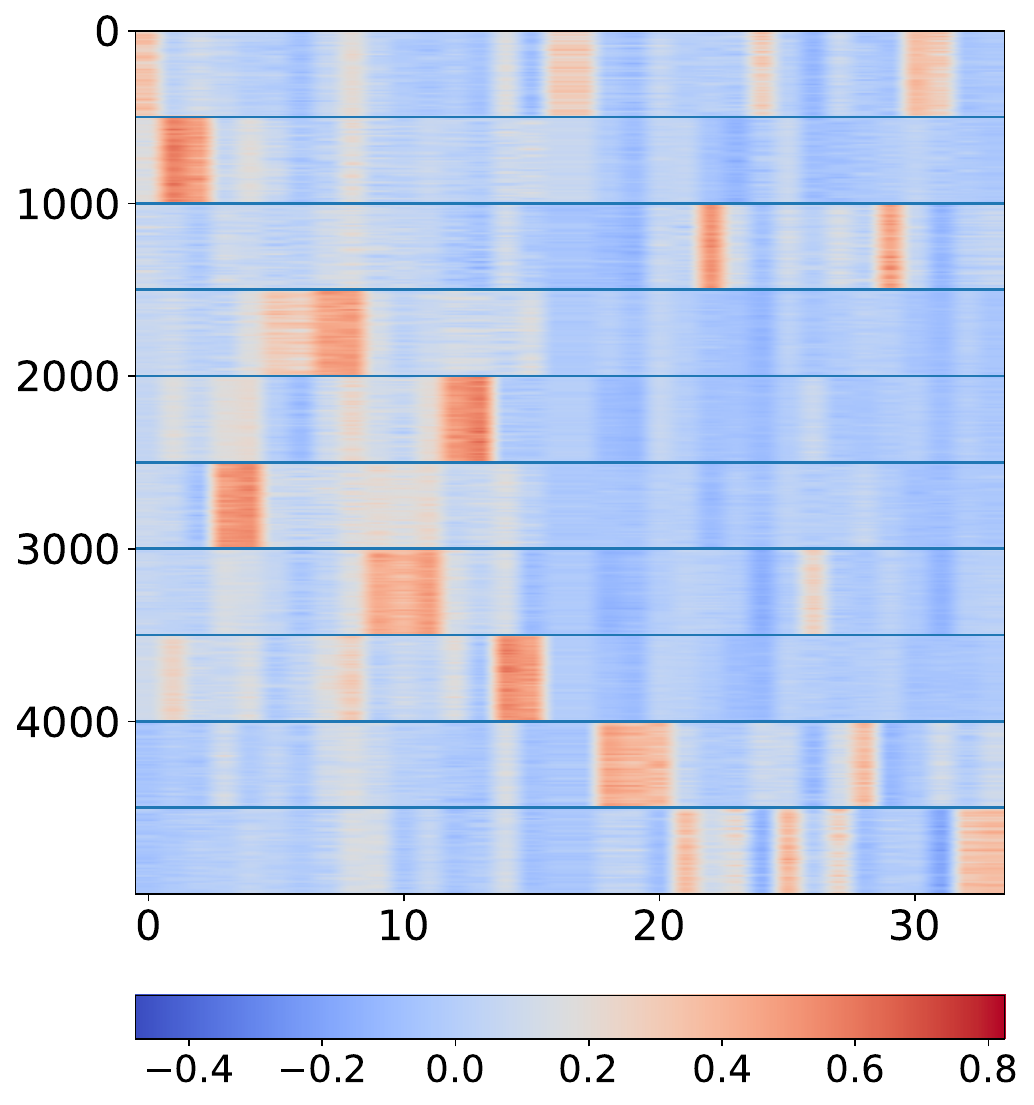}
        \caption{STL-10}
        \label{fig:image1}
    \end{subfigure}
    \hspace{0cm}  
    \begin{subfigure}[b]{0.49\columnwidth}
        \centering
        \includegraphics[width=\linewidth]{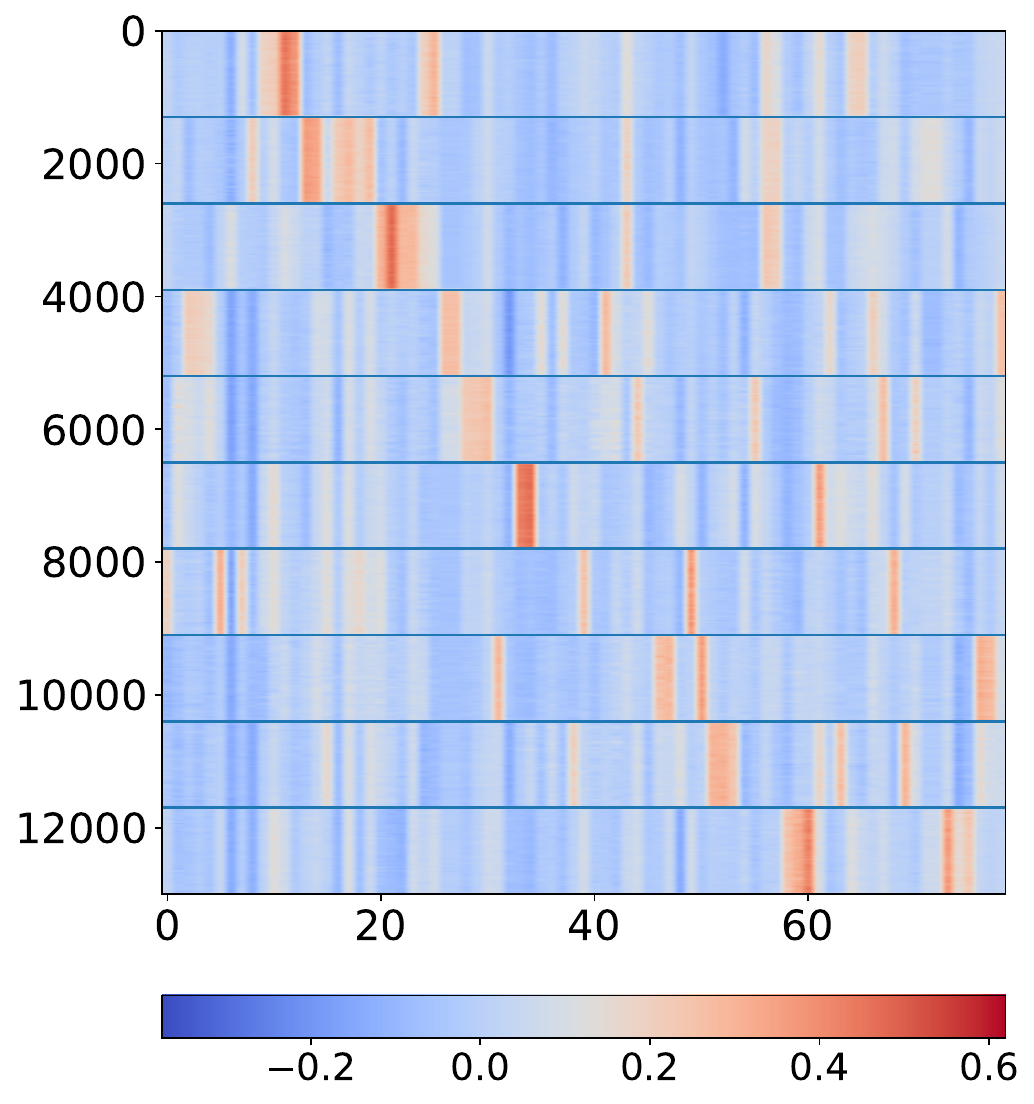}
        \caption{\normalfont ImageNet-10}
        \label{fig:image2}
    \end{subfigure}
    \caption{Visualization of image-text representation matrix $\mathbf{C}$.}
    \label{fig:C}
\end{figure}

\paragraph{Visualization of image-text representation matrix. }
 We extract cross-modal relational signals by learning an image-text representation matrix $\mathbf{C}\in\mathbb{R}^{N\times M}$ in Sec.~\ref{sec:3.1}. Each row $\mathbf{c}_i$ is associated with the image $x_i$ and serves as a new representation of it. To better demonstrate that these new representations show intra-class similarity and inter-class discriminability, we visualized the entire C matrices on the STL-10 and ImageNet-10 in Fig.~\ref{fig:C}. Rows associated with images of the same ground-truth class are grouped between two adjacent horizontal lines. 
It can be observed that within each class, the high and low value regions of different rows show clear consistency, indicating strong intra-class similarity; between different classes, the high and low value regions exhibit distinct differences, reflecting inter-class discriminability. The results are consistent with our analysis in Sec.~\ref{sec:3.1}.

\paragraph{Interpretability of learned semantic centers. }
To better illustrate the interpretability of the learned semantic centers in Sec.~\ref{sec:3.2}, we selected the most similar nouns of each learned semantic centers from the candidate noun set. Table~\ref{tab:prompt_comparison} reports the comparison between the selected nouns and the ground- truth class names. As can be shown, the selectde nouns for learned semantic centers align well with the ground-truth semantic of the dataset, demonstrating strong interpretability. 
Therefore, these results demonstrate that our method not only achieves effective clustering but also has the ability of providing accurate semantic descriptions for each category, thereby improving the richness of the clustering outcomes.

\begin{table}[h]
\centering
\caption{Comparisons between most similar nouns for semantic centers and corresponding ground-truth (noted as GT) class names. }
\resizebox{\columnwidth}{!}
{
\begin{tabular}{c|c|c|c}
\toprule
\multicolumn{2}{c|}{ImageNet-10} &
\multicolumn{2}{c}{STL-10}\\
\midrule
Most similar noun (ours) & GT class name & Most similar noun (ours) & GT class name\\ \midrule 
\texttt{sports\_car} & \textbf{Sports Car}         & \texttt{draft\_horse} & \textbf{Horse}  \\ 
\texttt{airline}  & \textbf{Airliner}              & \texttt{tabby\_cat} & \textbf{Cat}\\
\texttt{soccer\_ball}  & \textbf{Soccer Ball}      & \texttt{container\_ship} & \textbf{Ship}\\ 
\texttt{snow\_leopard}  & \textbf{Snow Leopard}    & \texttt{sports\_car} &\textbf{Car} \\ 
\texttt{king\_penguin} & \textbf{King Penguin}     & \texttt{riflebird} & \textbf{Bird}\\ 
\texttt{navel\_orange}  & \textbf{Orange}          & \texttt{guenon\_monkey} & \textbf{Monkey}\\ 
\texttt{Maltese\_dog}  & \textbf{Maltese Dog}      & \texttt{whitetail\_deer} & \textbf{Deer}\\ 
\texttt{Blimp}  & \textbf{Airship}                 & \texttt{trucking\_rig}  & \textbf{Truck}\\ 
\texttt{trucking\_rig}  & \textbf{Trailer Truck}   & \texttt{multiengine\_airplane} & \textbf{Airplane}\\ 
\texttt{containership} &\textbf{Container Ship}    & \texttt{domestic\_dog} & \textbf{Dog} \\ 
\bottomrule
\end{tabular}
}

\label{tab:prompt_comparison}
\end{table}

\subsection{Ablation Study}

\paragraph{Loss terms. }
To evaluate the effectiveness of the three loss terms $\mathcal{L_{\text{sup}}}$, $\mathcal{L_{\text{con}}}$, and $\mathcal{L_{\text{ent}}}$, we evaluate the performance of our method using different combinaions with these losses. From the results in Table \ref{tab:ablation_loss}, we observe that the supervised loss $\mathcal{L_{\text{sup}}}$ is crucial for establishing a solid clustering foundation, as evidenced by the performance when using only this term. Meanwhile, the consistency loss $\mathcal{L_{\text{con}}}$ and the entropy loss $\mathcal{L_{\text{ent}}}$ further enhance clustering performance, but their effects are limited without the supervised loss. Combining all three losses achieves the best performance, confirming that each loss term contributes to the overall effectiveness of our method.

\begin{table}[h]
\centering
\setlength{\tabcolsep}{5pt}
\renewcommand{\arraystretch}{1.15}
\small
\caption{Ablation results with different loss combinations.}
\resizebox{\columnwidth}{!}{%
\begin{tabular}{ccc|ccc|ccc}
\toprule
\multirow{2}{*}{$\mathcal{L_{\text{sup}}}$} & \multirow{2}{*}{$\mathcal{L_{\text{con}}}$} & \multirow{2}{*}{$\mathcal{L_{\text{ent}}}$}
& \multicolumn{3}{c|}{ImageNet-Dogs}
& \multicolumn{3}{c}{DTD} \\
\cmidrule(lr){4-6}\cmidrule(lr){7-9}
& & & NMI & ACC & ARI & NMI & ACC & ARI \\
\midrule
\checkmark &      &                  & 82.4 & 81.7 & 75.2 & 60.5 &  51.8 & 34.1 \\
      & \checkmark &                 & 8.1 & 14.3 & 3.6 & 10.2 &  12.4 &  8.6 \\
      &      & \checkmark            & 9.1 & 14.6 & 5.7 & 13.5 & 16.2 &  9.7 \\
\checkmark & \checkmark &            & \underline{86.0} & \underline{87.6} & 81.7 & 62.6 & 51.9 &  34.3 \\
\checkmark &      & \checkmark       & 85.9 & 87.4 & \underline{82.1} & \underline{63.3} & \underline{53.2} & \underline{35.5} \\
      & \checkmark & \checkmark      & 10.6 & 17.6 & 6.8 &  15.5 & 30.1 & 10.5\\
\checkmark & \checkmark & \checkmark & \textbf{86.2} & \textbf{88.0} & \textbf{82.6} & \textbf{63.8} & \textbf{53.7} & \textbf{36.2} \\
\bottomrule
\end{tabular}
}

\label{tab:ablation_loss}
\end{table}

\begin{figure*}[t]
    \centering
    \captionsetup[subfigure]{skip=0pt}
    \begin{subfigure}[t]{0.24\textwidth}
        \includegraphics[width=\linewidth]{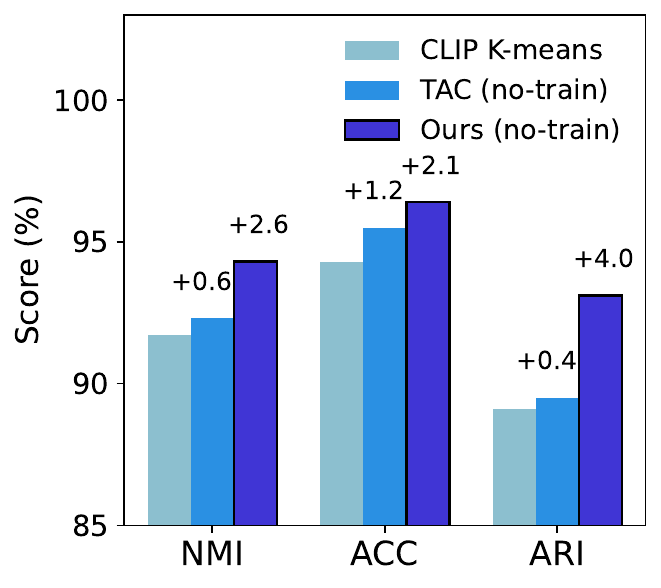}
        \caption{STL-10}
    \end{subfigure}\hspace{0pt}
    \begin{subfigure}[t]{0.242\textwidth}
        \includegraphics[width=\linewidth]{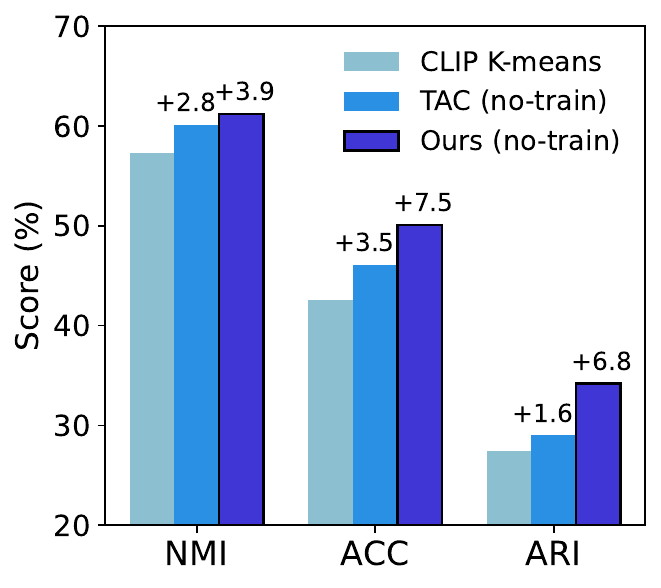}
        \caption{DTD}
    \end{subfigure}\hspace{0pt}
    \begin{subfigure}[t]{0.238\textwidth}
        \includegraphics[width=\linewidth]{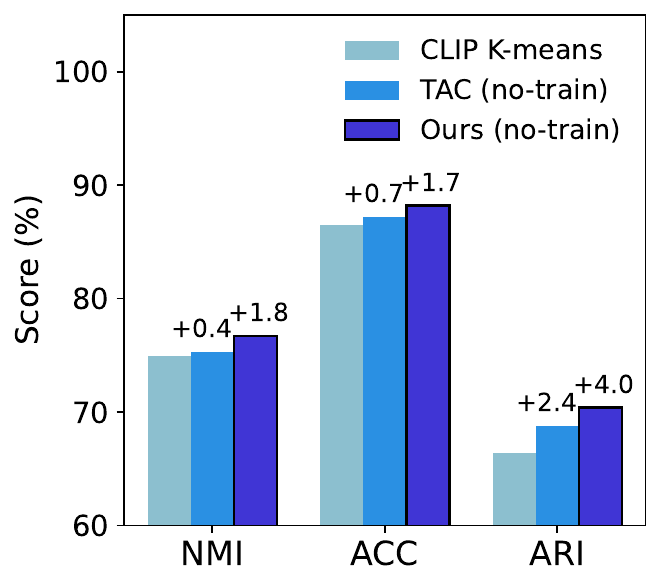}
        \caption{Flowers-102}
    \end{subfigure}\hspace{0pt}
    \begin{subfigure}[t]{0.24\textwidth}
        \includegraphics[width=\linewidth]{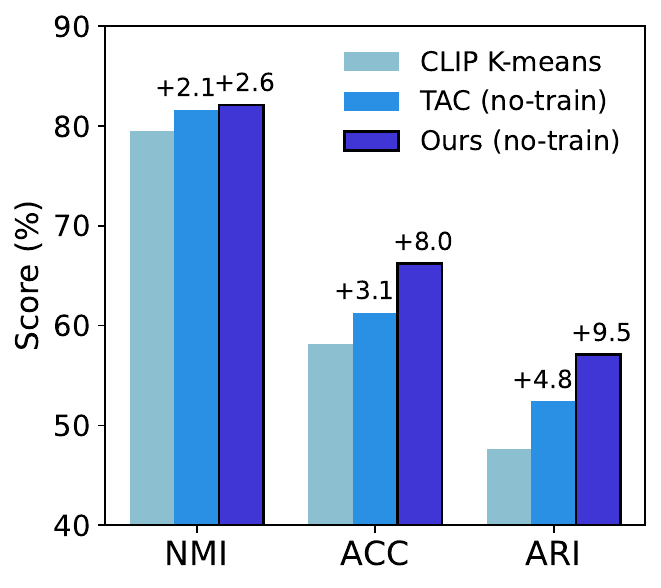}
        \caption{UCF-101}
    \end{subfigure}

    \caption{K-means performances comparisons on our method (Ours (no-train)) and two different baseline methods (CLIP K-means and TAC (no-train))across four datasets.}
    \label{fig:kmeans baselines}
\end{figure*}

\paragraph{Comparisons of different K-means baselines.} To demonstrate that we have extracted more discriminative relational signals in the first step, we directly perform K-means clustering on the image-text representation matrix $\mathbf{C}$ as a clustering baseline, referred to as Ours (no-train). We compare the performance of Ours (no-train) with two baseline methods, shown in Fig.~\ref{fig:kmeans baselines}. First, the CLIP baseline, which conducts K-means on image features from CLIP. Second, the TAC (no-train) baseline, which conducts K-means on features obtained by concatenating the images and their text counterparts. Compared to TAC (no-train), our method shows a greater improvement over the CLIP baseline. The improvement over both baselines is most notable on the DTD dataset (+4.0\% in ACC, +5.2\% in ARI compared to TAC (no-train) and +7.5\% in ACC, +6.8\% in ARI compared to CLIP baseline).

\begin{figure}[h]
    \centering
    \includegraphics[width=0.8\linewidth]{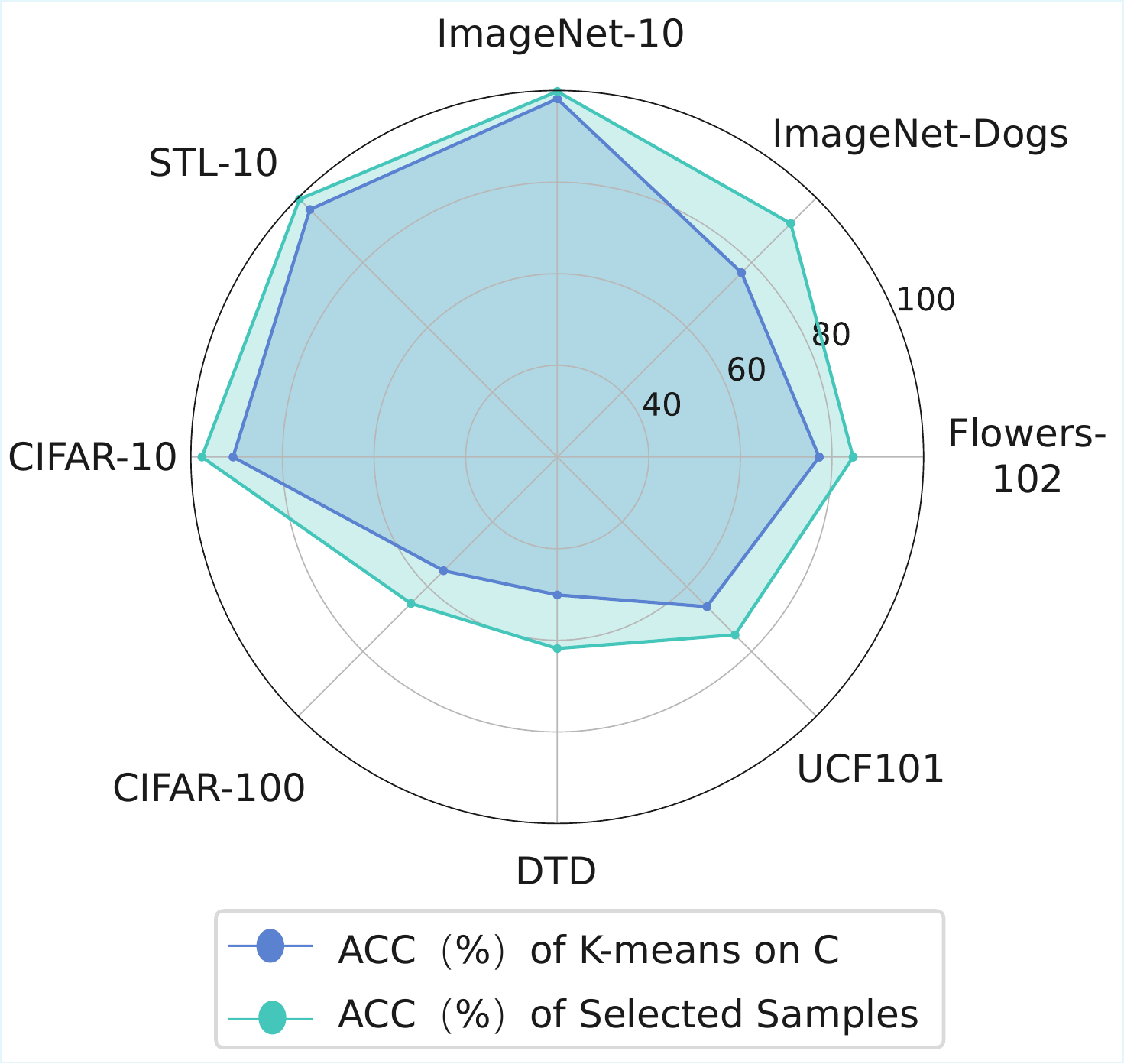}
    \caption{Accuracy gains achieved through neighborhood consistency filtering on seven datasets. Our strategy effectively mitigates clustering noise and ensures reliable supervision.}
    \label{fig:knn}
\end{figure}

\paragraph{Effectiveness analysis of selecting high-quality pseudo-labels. }
We compare the accuracy between pseudo-labels by conducting K-means clustering on $\textbf{C}$ and high-quality pseudo-labels after selecting. Results shown in Fig.~\ref{fig:knn} across eight diverse datasets demonstrate a substantial enhancement in pseudo-label accuracy following the neighborhood consistency filtering. Notably, significant gains are observed in datasets with lower performance on $\mathbf{C}$, such as ImageNet-Dogs (+15.2\%), CIFAR-100 (+10.1\%), and DTD (+11.7\%). These improvements validate that local structural consistency serves as a robust instruction for identifying high-quality samples while effectively decreasing noise near cluster boundaries. Thus, selection of high-quality subset provides a cleaner supervisory instruction for the learning of semantic centers.

\section{Conclusion}

In this paper, we propose a novel Language-Assisted Image Clustering (LAIC) framework that more effectively exploits the semantic knowledge embedded in VLMs. By revisiting the roles of text modality in existing LAIC methods, we observe two key limitations, including the weak discriminability of constructed text features and the restricted use of manually defined image–text semantic spaces during clustering. To overcome these issues, we introduce a cross-modal relation mining strategy that constructs an image–text representation matrix, enabling the extraction of more discriminative and compact representations. This representation allows effective clustering without additional supervision, providing a strong baseline via simple K-means clustering.
Furthermore, we propose a category-wise prompt learning scheme that learns continuous semantic centers from the native semantic space of the VLMs. By optimizing these centers to align with image features, our method goes beyond fixed image-text alignments and achieves more accurate and flexible semantic representations for clustering. Notably, the learned semantic centers not only improve clustering performance but also exhibit strong interpretability.
Experimental results show that our method outperforms state-of-the-art methods on eight benchmark datasets. This work highlights the better utilization of both modalities from VLMs for unsupervised clustering, providing insights for enhancing the generalization ability of VLMs on downstream tasks.

\section*{Impact Statement}
This paper presents work whose goal is to advance the field of language-assisted image clustering. There are many potential societal consequences of our work, none which we feel must be specifically highlighted here.


\bibliography{example_paper}
\bibliographystyle{icml2026}

\newpage
\appendix
\onecolumn


\section{Datasets Description}
\label{appendix datasets}
We conduct experiments on datasets from general to specialized and from simple to complex. First, four benchmark datasets for general classification are included. ImageNet-Dogs~\cite{IN-sub} is a fine-grained subset focused on dog breeds from ImageNet~\cite{in}. ImageNet-10~\cite{IN-sub} consists of ten broadly distinct general categories selected from ImageNet~\cite{in}. STL-10~\cite{STL10} is a dataset of ten common object categories. CIFAR-10~\cite{CIFAR} serves as the classic benchmark for general object classification at a low resolution of 32×32 pixels. In addition, four datasets with larger numbers of classes are covered. Flowers-102~\cite{flowers2}, which focuses on fine-grained classification of 102 flower species. CIFAR-100~\cite{CIFAR} extending the CIFAR-style framework to 100 fine-grained categories. DTD~\cite{DTD} includes 47 describable textures without reliance on object shapes. UCF-101~\cite{UCF} is a collection of realistic short video clips featuring 101 human action categories, used for video-based action recognition. The samples and number of classes information of all datasets used in our evaluation is summarized in Table ~\ref{table dataset}.
\begin{table}[h]
\caption{A summary of benchmark datasets used for evaluation.}
\label{table dataset}
\centering
{\small
\renewcommand{\arraystretch}{1.1}
\begin{tabular}{l|ccc}
\toprule
Dataset       & \#Samples & \#Classes             \\
\midrule
ImageNet-Dogs~\cite{IN-sub} & 19,500    & 15          \\
ImageNet-10~\cite{IN-sub}   & 13,000    & 10         \\
STL-10~\cite{STL10}        & 13,000    & 10         \\
CIFAR-10~\cite{CIFAR}      & 60,000    & 10       \\
\midrule
Flowers-102~\cite{flowers2}   & 8189  & 102      \\
CIFAR-100~\cite{CIFAR}     & 60,000    & 100      \\
DTD~\cite{DTD}           & 5,640     & 47          \\
UCF-101~\cite{UCF}        & 13,320    & 101        \\
\bottomrule
\end{tabular}
}
\end{table}

\section{Experimental Results under Multiple Random Seeds}
\label{appendix std}
To further evaluate the stability of our method, we report the average experimental results and standard deviation of our method and the comparison methods over 5 runs with different random seeds. The experiments are conducted on all datasets in Sec.~\ref{sec:4.1} and are shown in Table~\ref{table main result std} and Table~\ref{table main result 2 std}. It can be concluded that our method consistently improves clustering performance.

\begin{table}[h]
\centering
\caption{Average clustering performance (in percent \%) across four widely used benchmarks datasets.}
\label{table main result std}
\vskip 0.1in
\begin{small}
\resizebox{\textwidth}{!}{
\setlength{\tabcolsep}{3pt}
\begin{tabular}{l|ccc|ccc|ccc|ccc}
\toprule
\multicolumn{1}{l|}{Datasets} 
& \multicolumn{3}{c|}{ImageNet-Dogs}
& \multicolumn{3}{c|}{ImageNet-10} 
& \multicolumn{3}{c|}{STL-10}
& \multicolumn{3}{c}{CIFAR-10}\\
\midrule
Methods 
& NMI & ACC & ARI 
& NMI & ACC & ARI 
& NMI & ACC & ARI  
& NMI & ACC & ARI \\
\midrule


SIC \cite{SIC2023}
& 69.0$\pm$1.6 & 69.7$\pm$1.1 & 55.8$\pm$1.5
& 96.9$\pm$0.2 & 98.3$\pm$0.1 & 96.2$\pm$0.2 
& 95.3$\pm$0.1 & 98.1$\pm$0.1 & 95.9$\pm$0.1 
& 84.7$\pm$0.1 & 92.6$\pm$0.1 & 84.4$\pm$0.1 \\

MCA \cite{MCA2024} 
& 73.3$\pm$1.5 & 74.9$\pm$2.5 & 61.6$\pm$2.5
& - & - & - 
& 95.5$\pm$0.1 & 98.1$\pm$0.1 & 96.0$\pm$0.1 
& 84.9$\pm$0.2 & 92.7$\pm$0.2 & 84.6$\pm$0.2 \\

TAC \cite{TAC2024} 
& 80.9$\pm$1.0 & 83.3$\pm$1.4 & 72.0$\pm$1.0
& 98.2$\pm$0.1 & 99.2$\pm$0.1 & 98.3$\pm$0.1 
& 95.6$\pm$0.2 & 98.3$\pm$0.1 & 96.1$\pm$0.1 
& 83.4$\pm$0.2 & 92.1$\pm$0.2 & 83.4$\pm$0.3 \\

LFSS \cite{LFSS2025}
& 61.9$\pm$1.9 & 69.3$\pm$1.7  & 53.1$\pm$1.6
& 85.2$\pm$0.6 & 93.8$\pm$0.5 & 85.4$\pm$0.8 
& 77.3$\pm$1.1 & 86.1$\pm$0.9 & 74.0$\pm$0.8 
& 84.4$\pm$0.6 & 92.2$\pm$0.4 & 84.6$\pm$0.5 \\

PRO-DSC \cite{PRODSC2025}
& - & - & - 
& 98.0$\pm$0.3 & 99.0$\pm$0.2 & 97.8$\pm$0.2 
& 95.4$\pm$0.2 & 98.1$\pm$0.4 & 95.9$\pm$0.2 
& 79.6$\pm$0.4 & 87.1$\pm$0.3 & 80.2$\pm$0.4 \\


Ours
& \textbf{86.2}$\pm$1.1 & \textbf{88.0}$\pm$1.2 & \textbf{82.6}$\pm$1.2
& \textbf{99.6}$\pm$0.1 & \textbf{99.8}$\pm$0.1 & \textbf{99.7}$\pm$0.1 
& \textbf{96.1}$\pm$0.2 & \textbf{98.5}$\pm$0.1 & \textbf{96.7}$\pm$0.1 
& \textbf{85.2}$\pm$0.3 & \textbf{92.9}$\pm$0.3 & \textbf{84.8}$\pm$0.2 \\
\bottomrule
\end{tabular}
}
\end{small}
\vskip -0.1in
\end{table}

\begin{table}[h]
\centering
\caption{Average clustering performance (in percent \%) across four benchmarks datasets with larger cluster numbers.}
\label{table main result 2 std}
\vskip 0.1in
\begin{small}
\resizebox{\textwidth}{!}{
\setlength{\tabcolsep}{3pt}
\begin{tabular}{l|ccc|ccc|ccc|ccc}
\toprule
\multicolumn{1}{l|}{Datasets} & 
\multicolumn{3}{c|}{Flowers-102} &
\multicolumn{3}{c|}{CIFAR-100} & 
\multicolumn{3}{c|}{DTD} & 
\multicolumn{3}{c}{UCF-101} \\
\midrule
Methods &NMI &ACC &ARI &NMI &ACC &ARI &NMI &ACC &ARI &NMI &ACC &ARI \\
\midrule

SIC \cite{SIC2023} 
& 86.9$\pm$0.9 & 70.2$\pm$0.8 & 64.2$\pm$0.6
& 63.5$\pm$0.8 & 48.3$\pm$0.7 & 34.7$\pm$0.9
& 59.9$\pm$0.8 & 45.5$\pm$0.8 & 30.8$\pm$0.9 
& 81.7$\pm$0.8 & 62.1$\pm$1.1 & 52.5$\pm$0.9 \\
TAC \cite{TAC2024}
& 87.7$\pm$0.8 & 70.8$\pm$0.9 & 63.9$\pm$0.5
& 68.1$\pm$0.2 & 55.4$\pm$0.7 & 42.4$\pm$0.6
& 61.8$\pm$0.5 & 50.9$\pm$0.6 & 34.8$\pm$0.7 
& 82.5$\pm$0.4 & 69.1$\pm$0.8 & 60.2$\pm$0.7\\
LFSS \cite{LFSS2025} 
& 79.2$\pm$1.1 & 55.4$\pm$1.2 & 51.8$\pm$0.9
& 66.0$\pm$0.6 & 52.4$\pm$0.8 & 37.6$\pm$0.8
& 58.7$\pm$1.1 & 45.5$\pm$1.2 & 31.9$\pm$0.9
& 58.9$\pm$0.9 & 34.5$\pm$1.2 & 21.6$\pm$1.1\\
PRO-DSC \cite{PRODSC2025} 
& 83.2$\pm$1.0 & 69.8$\pm$0.8 & 59.6$\pm$0.7
& 68.2$\pm$0.5 & 55.7$\pm$0.6 & 40.2$\pm$0.3  
& 57.6$\pm$0.7 & 48.1$\pm$0.9 & 31.3$\pm$0.8 
& 81.0$\pm$0.8 & 61.9$\pm$0.9 & 52.8$\pm$0.7 \\
Ours
& \textbf{89.6}$\pm$0.8 & \textbf{78.6}$\pm$0.7 & \textbf{72.9}$\pm$0.4
& \textbf{68.8}$\pm$0.2 & \textbf{58.1}$\pm$0.4 & \textbf{43.4}$\pm$0.3
& \textbf{63.8}$\pm$0.6 & \textbf{53.7}$\pm$0.6 & \textbf{36.2}$\pm$0.5 
& \textbf{84.4}$\pm$0.5 & \textbf{70.2}$\pm$0.7 &\textbf{61.9}$\pm$0.9\\

\bottomrule
\end{tabular}
}
\end{small}
\vskip -0.1in
\end{table}

\section{Hyper-parameter Analyses}

\begin{figure}
    \centering
    \includegraphics[width=0.8\linewidth]{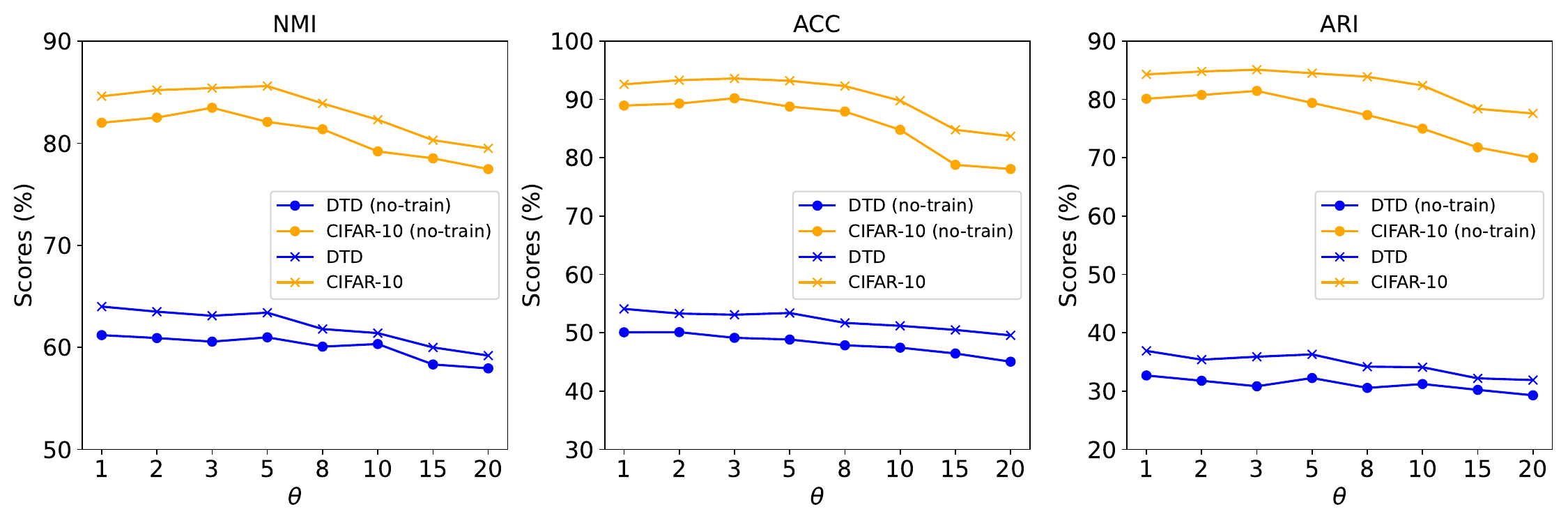}
    \caption{Clustering performances with different $\theta$ on CIFAR-10 and DTD.}
    \label{fig:placeholder}
\end{figure}
\begin{figure}
    \centering
    \includegraphics[width=0.8\linewidth]{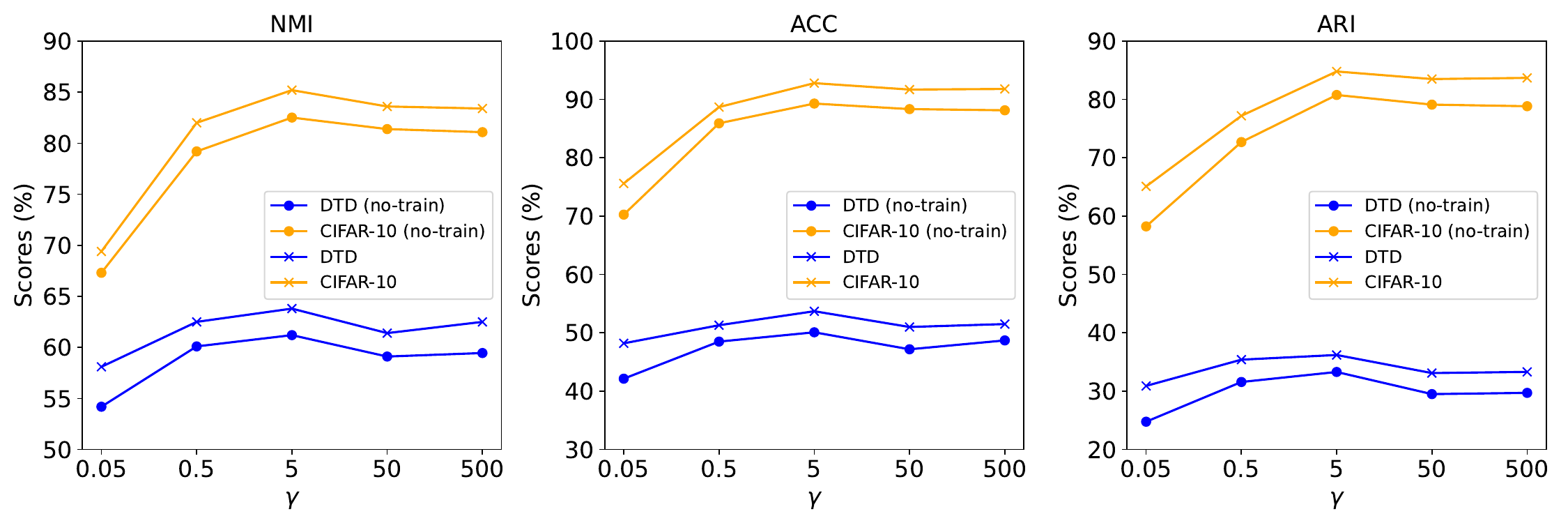}
    \caption{Clustering performances with different $\gamma$ on CIFAR-10 and DTD.}
    \label{fig:k}
\end{figure}

\paragraph{Number of closed nouns selected for fine-grained image center $\theta$.}
In Sec.~\ref{sec:3.1}, to construct a candidate noun set that describes the images in the current dataset, we first obtain fine-grained image centers and assign $\theta$ nouns to each center to represent its semantics. To investigate the effect of $\theta$, we evaluate the performance of directly applying K-means on $\mathbf{C}$ (noted as ``no-train") and the final clustering performance under different values of $\theta$ on CIFAR-10 and DTD.
We observe that when $\theta$ is small, the clustering performance remains relatively stable, while it degrades noticeably as $\theta$ increases. We think that a small $\theta$ allows the selected candidate nouns to accurately capture the semantic of different classes. In this case, nouns that are close to a given image provide reliable semantic descriptions. Meanwhile, distant nouns are likely associated with other classes. They can provide meaningful discriminative information, indicating that the image is not close to a certain other class. However, when $\theta$ becomes large, the candidate noun set tends to include more nouns that are irrelevant to the dataset, which affects the quality of the $\mathbf{C}$ .
Based on these observations, we set $\theta = 2$ in all experiments.

\paragraph{Regularization parameter $\gamma$. }
In Problem~\eqref{eq:ridge}, we use Frobenius regularizer to impose a smoothness constraint on $\mathbf{C}$, preventing it from being overly sparse, and $\gamma$ is the regularization parameter.
We obtain the matrix $\mathbf{C}$ under different values of $\gamma$ on CIFAR-10 and DTD, and report both the clustering results obtained by directly applying K-means on $\mathbf{C}$ (noted as ``no-train") and the final clustering performance (shown in Fig.~\ref{fig:k}).
We observe that the clustering performance degrades noticeably when $\gamma$ is smaller than 5, while it remains relatively stable as $\gamma$ is set larger.
When $\gamma$ is overly small, the matrix becomes much sparse, which limits the number of text features involved in describing the image, resulting in worse discriminability. This suggests that describing the image from the global candidate noun set is appropriate. When $\gamma$ is too large, the matrix becomes overly smooth, with each row becoming more similar and introducing redundant information, and also reduce the discriminability of $\textbf{C}$.
Our method is robust to the choice of $\gamma$ within a reasonably large range, and is primarily sensitive only to overly small values of the regularization parameter.
We fix $\gamma = 5$ in all experiments to achieve a favorable balance between reconstruction accuracy and richness.
\begin{figure}
    \centering
    \includegraphics[width=1\linewidth]{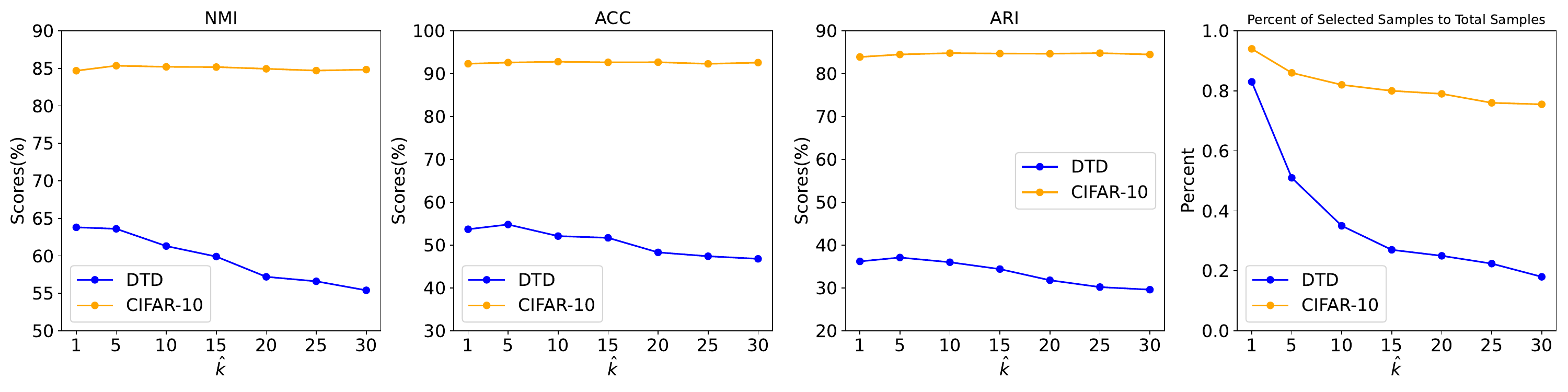}
    \caption{Clustering performances and percent of selected samples to total samples with different $\hat{k}$ on CIFAR-10 and DTD.}
    \label{fig:gamma}
\end{figure}
\paragraph{Number of nearest neighbors $\hat{k}$. }
In Sec.~\ref{sec:3.2}, we filter high-quality samples by measuring the consistency between each sample and its $\hat{k}$ nearest neighbors. We evaluate the clustering results on CIFAR-10 and DTD under different choices of $\hat{k}$, reporting ACC, NMI, and ARI, as well as the percentage of selected high-quality samples with respect to the total number of samples (shown in Fig.~\ref{fig:gamma}).
We observe that even with a large $\hat{k}$, a sufficient number of high-quality samples can still be selected on CIFAR-10, making the final clustering performance relatively insensitive to the choice of $\hat{k}$. In contrast, the DTD dataset has a much larger clustering number of. As $\hat{k}$ increases, the requirement that a sample and all its $\hat{k}$ nearest neighbors share the same pseudo-label becomes overly strict. This leads to an obvious reduction in the number of selected samples and degrades clustering performance.
Thus, for datasets with a smaller number of clusters, the choice of $\hat{k}$ can be relatively flexible, which we set to $10$. For datasets with a larger number of clusters, we set $\hat{k}=1$ to balance both the quantity and quality of the selected samples.

\section{Experiments on More Backbones}
We validate the effectiveness of extracting supervision signals through cross-modal relations on more VLMs. We selected the ViT-B/16 architecture of the BLIP~\cite{blip}, constructed image and candidate noun sets using the same strategy as in Sec.~\ref{sec:3.1}, and compared the TAC (no-train) baseline with Ours (no-train) baseline. Experiments on three datasets show that our supervision signal extraction strategy remains effective with BLIP's image-text features, demonstrating the generalizability of our observations and method.
\begin{table}[h]
\centering
\caption{Comparisons with various methods on clustering performance (in percent \%) across three benchmarks datasets.}
\label{table blip}

\begin{small}
\setlength{\tabcolsep}{3pt}
\begin{tabular}{l|c|ccc|ccc|ccc}
\toprule
\multicolumn{2}{l|}{} & 
\multicolumn{3}{c|}{CIFAR-10} &
\multicolumn{3}{c|}{DTD-47} & 
\multicolumn{3}{c}{UCF-101} \\
\midrule
Methods & Backbone &NMI &ACC &ARI &NMI &ACC &ARI &NMI &ACC &ARI  \\
\midrule
TAC (no-train) \cite{TAC2024} & BLIP ViT-B/16
& 79.9 & 88.9 & 77.5
& 62.3 & 52.1 & 35.9 
& 76.1 & 60.9 & 51.4\\
Ours (no-train) & BLIP ViT-B/16
&\textbf{ 83.6} & \textbf{92.1} &\textbf{ 83.5}
& \textbf{65.1} & \textbf{57.2 }& \textbf{41.7 }
&\textbf{ 78.9} & \textbf{62.9} & \textbf{53.6}\\


\bottomrule
\end{tabular}
\end{small}

\end{table}

\end{document}